\documentclass[ journal]{IEEEtran}

\usepackage{amsmath}
\usepackage{amssymb}
\usepackage{graphicx}
\graphicspath{{img/}}
\usepackage{booktabs,colortbl}
\definecolor{Gray}{gray}{0.9}
\usepackage{multirow}
\usepackage{times}
\usepackage{mathptmx}
\usepackage[caption = false]{subfig}
\usepackage{comment}
\usepackage{balance}
\usepackage{standalone}

\makeatletter
\newcommand*{\addFileDependency}[1]{
  \typeout{(#1)}
  \@addtofilelist{#1}
  \IfFileExists{#1}{}{\typeout{No file #1.}}
}
\makeatother

\DeclareMathOperator{\logit}{logit}
\newcommand{\pmo}{p_{\text{mort}}}

\begin{document}
\title{
  \color{black}
  A Large-scale Multimodal Study for Predicting Mortality Risk 
  Using Minimal and Low Parameter Models  and Separable Risk Assessment
  \color{black}
}

\author{Alvaro E. Ulloa Cerna,~\IEEEmembership{Member, ~IEEE}, 
    David P. vanMaanen,
    Linyuan Jing, 
    Joshua V. Stough,
    Aalpen A. Patel,\\
    Christopher M. Haggerty,
    Brandon K. Fornwalt, and
    Marios S. Pattichis~\IEEEmembership{Senior Member, ~IEEE}
    \thanks{This project was funded, in part, under a grant with the Pennsylvania Department of Health (\#SAP 4100070267).}%
    \thanks{Corresponding author: MP, Email: pattichi@unm.edu}
    \thanks{AUC, LJ, CH and BF e-mails: [alvarouc, jinglinyuan2008, chris.m.haggerty, brandon.fornwalt]@gmail.com, JV e-mail: joshua.stough@bucknell.edu, DV email: david@vanmaanen.us, AP email: aapatel@geisinger.edu}
    \thanks{During the time of the study, AUC, DV, LJ, AP, JV, CH, and BF were with the Department of Translational Data Science and Informatics, Geisinger, PA 17822, USA. AUC and MP are with the Department of Electrical and Computer Engineering, University of New Mexico, NM 87131, USA.}}
    
\markboth{Journal of Biomedical and Health Informatics}{Ulloa \MakeLowercase{\textit{et al.}}:
 A Large-scale Multimodal Study for Predicting Mortality Risk}
\IEEEtitleabstractindextext{%
  \begin{abstract}
     \color{black}   
    The majority of biomedical studies use limited datasets
        that may not generalize over large heterogeneous datasets that
        have been collected over several decades.
    The current paper develops and validates several multimodal models
        that can predict 1-year mortality based on a massive clinical dataset.
    Our focus on predicting 1-year mortality 
         can provide a sense of urgency to the patients. 
           
    Using the largest dataset of its kind, the paper considers the development and 
        validation of multimodal models based on 25,137,015 videos 
        associated with 699,822 echocardiography studies from 316,125 patients, 
        and 2,922,990 8-lead electrocardiogram (ECG) traces from 631,353 patients. 
    Our models allow us to assess the 
       contribution of individual factors and modalities to the overall risk. 
    Our approach allows us to develop extremely low-parameter models
          that use optimized feature selection based on feature importance.
    Based on available clinical information, we construct a family of models
         that are made available in the DISIML package.
    Overall, performance
         ranges from an AUC of 0.72 with just ten parameters to 
         an AUC of 0.89 with under 105k for the full multimodal model.
    \color{black}
    The proposed approach represents a 
         modular neural network framework that can provide insights into 
         global risk trends and guide therapies for reducing mortality risk. 
\end{abstract}
}
\maketitle

\IEEEdisplaynontitleabstractindextext

\section{Introduction}
\label{sec:introduction}
The overwhelming majority of biomedical image and video analysis methods are
    evaluated on very limited datasets.
With the evolution of Deep Learning systems with large parameters,
    the standard approach is to use transfer learning before
    retraining the models for application on biomedical datasets.
Unfortunately, with the use of transfer learning, it is not clear what the derived models learn
    from unrelated datasets.
The use of large biomedical datasets allows us to
    train relatively low-parameter models on unique
    biomedical tasks without the need for transfer learning.

The successful adoption of electronic health records (EHR) 
    has made it possible to design studies with large multimodal datasets.
Furthermore, the use of large multimodal datasets can be used
    to develop generalizable models that can support
    clinical decision making through explanations
    supported by multiple modalities.
Here, we want to consider laboratory measurements,
      ECG or other 1D signals, as well as
      imaging data.

Multimodal risk models based on large datasets
    can clearly benefit precision medicine (see \cite{foley2016european, allen2017use, Precision}).
On the other hand, there is a clear need for the models to support interpretability (see \cite{bonner2018clinical})
    that includes all relevant predictors 
    (or, conversely, clear explanations when relevant data input is excluded)~\cite{kattan2016american}. 
Clinical decisions need to be explained in terms of the factors that influenced the decision
    as required by the European General Data Protection Regulation (https://eugdpr.org/).
In the current paper, we develop a separable risk model that can be used
    to quantify how specific factors contribute to the overall risk.

\color{black}
We note that risk models are routinely used by physicians in clinical practice.
These models do not predict time to death.
Our risk models assess risk within a fixed period.      
Similarly, the Framingham risk score~\cite{wilson1998prediction}
     provides a score for the risk of developing atherosclerotic 
     cardiovascular disease or events within ten years.
Other examples include     
    the Pooled Cohort Equations~\cite{yadlowsky2018clinical} 
    for atherosclerotic cardiovascular disease,  and the Seattle Heart 
    Failure score~\cite{levy2006seattle},
    for predicting mortality for 1, 2, and 3 years
    for patients who have suffered heart failure.

The current paper assesses mortality risk within a year.
Thus, there is a natural urgency to the prediction since patients
       are much more likely to react to short-term rather than long-term risks.
The current paper presents the first large-scale multimodal study focused
    on one-year mortality risk prediction based on EHR tabular data,
    echocardiography views, and ECG data.
\color{black}    
Our datasets are drawn from 25,137,015 videos in 699,822 echocardiography studies from 316,125 patients, and
    2,922,990 8-lead ECG traces from 631,353 patients.
Derived from this dataset, we consider a total
    of 28 combinations with three different methods for a total of
    28*3=84 models for predicting one-year mortality.
We are not aware of any published study that is comparable to this dataset.
In general, previously published research uses a limited number of models
   on much smaller datasets.

Prior research on the use of large-scale video datasets has been 
   primarily limited to echocardiography studies.
In  \cite{ouyang2020video}, the authors introduced the EchoNet-Dynamic, 
     a video based system using 3D CNN and residual connections, for the assessment 
    of cardiac function using a relatively large dataset of  10,030 annotated echocardiography videos.
EchoNet-Dynamic performed very well at
   segmentating the left ventricle, 
   estimating the ejection fraction,
   and then assessing heart failure due to a reduced ejection fraction.
The authors verified that the system performance variance was
   comparable to what can be achieved with human observers.
We note that EchoNet-Dynamic used the standard apical four-chamber
    view and also used weak supervision from expert human tracings
    (see Fig. 1 caption of \cite{ouyang2020video}).
     
In \cite{duffy2022high}, the authors used the EchoNet-Dynamic system
      to measure left ventricular wall thickness on
      23,745 patients based on parasternal long-axis or apical 4-chamber view echocardiography videos. 
The authors reported that the system was able to detect subtle changes 
     in left ventricular thickness that were then used to establish the causes of hypertrophy
     (e.g., cardiac amyloidosis or hypertrophic cardiomyopathy).

In  \cite{reddy2023video}, the authors fine-tuned the EchoNet-Dynamic system
     on a  pediatric data set of  1,958 patients
     with   4,467 2-D apical 4-chamber and parasternal short axis echocardiography views
     to calculate the left ventricular ejection fraction based on  select video frames.
EchoNet-Peds was trained on 80\% of the dataset and tested on the remaining 20\%
   and found that the system could be used to identify pediatric patients with systolic 
   dysfunction.

Compared to these earlier studies, we note that the current study
    is much larger, with data taken from 631,353 patients.
Furthermore, instead of two views,
    video analysis is performed on 6 echocardiography
    views without any requirement for human supervision.
Beyond the use of video models,
    our study  developed 84 low-parameter multimodal
    models that are used to predict 
    all-cause 1 year mortality as verified by electronic health records.

Our current paper builds on our prior research that 
    developed a limited number of prediction models.
In what follows, we provide a summary of our own related research.
In terms of models, we developed 
   one-year mortality risk models using only EHR tabular data in~\cite{samad2018predicting}, 
   electrocardiography (ECG) data with age and sex features in~\cite{raghunath2020prediction}, and 
    echocardiography with tabular data in~\cite{ulloa2020echo}. 
In~\cite{samad2018predicting}, we trained a Random Forest model with tabular information from 
   patient demographics, blood test panels, and measurements from echocardiography studies. The random forest model yielded an area under the receiver operating characteristics curve (AUC) of 0.85, and was cross-validated within a 171,510 patient dataset.
In~\cite{ulloa2020echo}, we considered the development of video models
    that were derived from 42,095 echocardiography studies, compared
    to the 699,822 echocardiography studies used in the current paper.
In~\cite{ulloa2020echo}, our findings suggested that video data 
   improved the performance of predicting 1-year mortality  
   and that the model would benefit from more samples. 
Specifically, classification performance as a function of 
   sample size did not reach a saturation point.
In~\cite{raghunath2020prediction}, we trained 1-D convolutional neural networks (CNN) 
  with raw trace data from ECG that yielded an AUC of 0.88 in a held-out dataset of 168,914 patients.
In~\cite{ulloa2020echo}, we trained 3-D CNNs with raw pixel data from echocardiography videos that 
   yielded an AUC of 0.83 using video data only and 0.84 combining video and tabular EHR data. The video models were cross-validated with a 34,362 patients dataset. 

\color{black}
The current study aims to develop a family of separable risk models
       that can be used to predict one-year mortality based on clinical 
       information that is available to us.
Our focus on one-year mortality provides urgency for
       potential implementation of our work in clinical practice.       
We consider a large sample size of 631,353 patients to develop and validate our models.
In terms of clinical information, we are using a multimodal dataset
       that includes echocardiography videos from multiple views,
       8-lead ECG traces, and EHR tabular data.
Our use of a multimodal dataset
     is significant because it represents
     a more complete picture of information that would be available
     for assessing mortality risk.
Then, we develop a large number of prediction models that consider
    all possible combinations of the different modalities.
For model assessment, we report AUC results from training 
     on the individual modalities
    as well as their combinations.
Given the size of our dataset, training required significant 
    data storage and computational resources.
It took approximately 45 days to train
    the models on an NVIDIA DGX-2 platform with 16 V100 32GB GPUs.
Our development of separable risk assessment models allow us to 
    quantify how different factors contribute 
    to the overall risk.
Specifically, our separable model
    allows us to assess the importance of each factor and
    provides an independent global assessment of the contribution of each
    factor.
Overall, we develop low-parameter
    models that are trained on our large-scale
    datasets while avoiding biases associated
    with transfer learning methods.
Among the many models considered here,
    we also build minimal models based on ranked factors.
We then compare performance across models to demonstrate
    how the use of more complex models achieves more accuracy
    while simpler models achieve less accuracy while being
    easier to interpret.
\color{black}    

The remainder of the paper is organized into four sections. In section \ref{sec:dataset}, we describe the
dataset. In section \ref{sec:methods}, we describe the proposed methodology. 
We then provide results and discussion in section
\ref{sec:results} and concluding remarks in section~\ref{sec:concl}.

\begin{table}     \label{tab:varlist}
\caption{List of variables for each data modality. Units are given in between parentheses. Binary variables do not have units.}
\begin{tabular}{p{.97\columnwidth}}
 \toprule
 Clinical EHR \\ \midrule
 \textbf{Demographics:}  White, 
 Male,
 Ever smoked,
 Age (years) \\ 
  \textbf{Vitals:}
 Body mass index (kg/m$^2$),
 Diastolic Pressure (mmHg),
 Systolic Pressure (mmHg),
 Heart Rate (beats/min),
 Height (cm),
 Weight (kg)\\
 \textbf{Labs:}
 HbA1C (\%),
 Bilirubin (mg/dL),
 BUN (mg/dL),
 Cholesterol (mg/dL),
 Creatine kinase-MB (ng/mL),
 Creatinine (mg/dL),
 C-reactive protein (mg/L),
 D-dimer ($\mu$g/mL),
 Glucose (mg/dL),
 High-density lipoprotein (mg/dL),
 Hemoglobin (g/dL),
 Lactate dehydrogenase (U/L),
 Low-density lipoprotein (mg/dL),
 Lymphocytes (\%),
 Potassium (mmol/L), 
 Pro-BNP (pg/mL),
 Sodium (mmol/L),
 Troponin I (ng/mL), 
 Troponin T (ng/mL), 
 Triglyceride (mg/dL),
 Uric acid (mg/dL),
 Very-low-density lipoprotein (mg/dL),
 eGFR (mL/(min 1.73 m$^2$)), \\
\midrule
Echocardiography \\ \midrule
\textbf{Aortic valve:} No regurgitation,
 Mild regurgitation,
 Moderate regurgitation,
 Severe regurgitation,
 Regurgitation not assessed,
 No stenosis,
 Mild stenosis,
 Moderate stenosis,
 Severe stenosis,
 Stenosis not assessed,
 Insufficiency deceleration slope (cm/sec$^2$),
 Insufficiency max velocity (cm/sec),
 Area (cm$^2$(I,A) ),
 Flow vel-time integral (cm),
 Flow distal max vel (cm/sec),
 Flow distal mean vel (cm/sec),
 Mean pressure gradient (mmHg)\\
\textbf{Mitral valve:} No regurgitation,
 Mild regurgitation,
 Moderate regurgitation,
 Severe regurgitation,
 Regurgitation not assessed,
 No stenosis,
 Mild stenosis,
 Moderate stenosis,
 Severe stenosis,
 Stenosis not assessed,
 Regurgitation max vel (cm/sec),
 A-point flow max vel (cm/sec),
 E-point flow max vel (cm/sec),
 P1/2t max vel (cm/sec),
 Deceleration slope (cm/sec$^2$),
 Deceleration time (sec)\\
\textbf{Pulmonary valve:}
No regurgitation,
 Mild regurgitation,
 Moderate regurgitation,
 Severe regurgitation,
 Regurgitation not assessed,
 No stenosis,
 Mild stenosis,
 Moderate stenosis,
 Severe stenosis,
 Stenosis not assessed,\\
 \textbf{Tricuspid valve:}
 No regurgitation,
 Mild regurgitation,
 Moderate regurgitation,
 Severe regurgitation,
 Regurgitation not assessed,
 No stenosis,
 Mild stenosis,
 Moderate stenosis,
 Severe stenosis,
 Stenosis not assessed,
 Regurgitation max vel (cm/sec), \\
 \textbf{Left ventricle:}
 End-diastolic vol (AP2)$^*$ (ml), 
 End-diastolic vol (AP4)$^*$ (ml),
 End-diastolic vol (AP2)$^{**}$ (ml),
 End-diastolic vol (AP4)$^{**}$ (ml),
 End-systolic vol (AP2)$^*$ (ml),
 End-systolic vol (AP4)$^*$ (ml),
 End-systolic vol (AP2)$^{**}$ (ml),
 End-systolic vol (AP4)$^{**}$ (ml),
 V1 vel-time integral, 
 V1 max vel (cm/sec), 
 V1 mean vel (cm/sec), 
 Diastole area (AP2) (cm$^2$),
 Diastole area (AP4) (cm$^2$), 
 Systole area (AP2) (cm$^2$), 
 Systole area (AP4) (cm$^2$), 
 Internal diastole dimension (cm),
 Internal systole dimension (cm),
 Diastole length (AP2) (cm),
 Diastole length (AP4) (cm), 
 Systole length (AP2) (cm), 
 Systole length (AP4) (cm), 
 Outflow tract area (cm$^2$),
 Outflow tract diameter (cm), 
 Diastole posterior wall thickness (mm),
 Ejection fraction (\%)\\
\textbf{Left atrium:}
 Left atrium dimension (cm),
 Left atrium vol (AP2)$^*$ (ml),
 Left atrium vol (AP4)$^*$ (ml) \\
\textbf{Pulmonary artery:}
V2 max vel (cm/sec), 
 Acceleration slope (cm/sec$^2$), 
 Acceleration time (sec), \\
\textbf{Other:}
 Aortic root diameter (cm), 
 Interventricular septum thickness (cm),
 Pulmonary R-R interval (sec), 
 Right atrial end-systolic mean pressure (mmHg),
 Right ventricle dimension at end-diastole,
 Ascending aorta diameter (cm), 
 Normal diastolic function,\\ \midrule
Electrocardiography (ECG)  \\ \midrule
\textbf{Findings:}
Acute myocardial infarction,
 Atrial fibrillation, 
 Atrial flutter,
 Complete heart block,
 Early repolarization,
 Fascicular block,
 First-degree block,
 Intraventricular block,
 Incomplete left bundle-branch block, 
 Incomplete right bundle-branch block, 
 Ischemia, 
 Left axis deviation, 
 Left bundle-branch block,
 Low QRS voltage, 
 Left ventricular hypertrophy,
 Nonspecific ST changes,
 Nonspecific T-wave changes,
 Normal, 
 Other bradycardia, 
 Premature atrial contractions, 
 Pacemaker,
 Poor tracing, 
 Previous infarct,
 Previous anterior myocardial infarction,
 Prolonged QT, 
 Premature ventricular contractions,
 Right axis deviation,
 Right bundle-branch block, 
 Second-degree atrioventricular block,
 Sinus bradycardia, 
 Supraventricular tachycardia,
 Tachycardia,
 T-wave inversion,
 Ventricular tachycardia, \\
  \textbf{Measurements:}
R axis (degrees),
 PR interval (sec),
 P axis (degrees), 
 QRS duration (sec), 
 QT (msec), 
 QTC (msec),
 T axis (degrees), 
 Ventricular rate (beats/min),
 Average RR interval (msec)\\ 
 \bottomrule
$^*$ Obtained by modified ellipsoid technique\\
$^{**}$ Obtained using a single plane ellipsoid technique\\
 \end{tabular}
\end{table}

\section{Large-scale Datasets}\label{sec:dataset}
\subsection{Electronic Health Records}\label{sec:ehr}
Geisinger's EHR database contained demographics, vitals, labs, and measurements and findings from ECG and echocardiography studies from a total of 2,135,458 patients, of which 733,245 patients had at least an ECG or an echocardiography study.
Geisinger's ECG database contained 2,922,990 ECGs from 631,353 patients acquired over 37 years (January 1984 to September 2021).
Geisinger's echocardiography database
contained 699,822 studies (25,137,015 videos) from 316,125 patients performed over
23 years (February 1998 to September 2021). 
Each study included the date, a findings report, and patient identifier information.
Geisinger's Phenomics Initiative database has modeled these data along 
   with multiple clinical features into a tabular format that included
   human-derived echocardiography and ECG measurements.
We then
   combined all datasets and merged them based on echocardiography encounters. 
We assigned the latest measurements up to a year prior to the  
   echocardiography encounter. If the measurement was older than 1 year, it was considered missing. 
This approach was also applied to the ECG dataset, where for every echocardiography encounter we linked a past ECG no older than a year.  


The measurements were cleaned from input errors 
    that had resulted in entries
    that were not physiologically possible.
In cases where we could not establish limits,
    we removed extreme outliers.
Here, we defined extreme outliers as values 
    that were more than three standard deviations from the mean,
    or below the 25th percentile minus 3 interquartile ranges,
    and similarly for the 75th percentile.
After removing the outlier values, we set the values as missing.

\color{black}
We applied a Multiple Imputation by Chained Equations  
     (MICE)~\cite{buuren2010mice}  for handling missing values.
\color{black}  
We note that the effect of replacing the missing values has
   a minimal effect on the model AUC \cite{farhangfar2008impact}, \cite{samad2018predicting}.

The diastolic function was classified as normal or dysfunction (I, II, III or unspecified).
We define our one year mortality risk based on the date of the echocardiography study.
To establish the patient status (dead/alive),
   we used last exam, or a confirmed death date within our national death databases.

By definition, patients in the ``Alive'' (negative) class were followed for over a 
   year (i.e., had a last-known living encounter more than one year after the echocardiography study), while patients in the ``Deceased'' (positive) class were followed for less than a year (until death). 
Thus, patients in the negative class have longer follow-up time compared to 
   patients in the positive class and contributed to more available echocardiography studies. 
The 539,689 studies that fell in the alive class belonged to 248,250 unique patients
   (2.2 studies per patient), and the 68,049 studies from the deceased class belonged to 46,031 patients (1.5 studies per patient). 
To avoid biasing the results to patients in the ``Alive'' class, 
   we report performance metrics after selecting a random study per patient.

\subsection{Echocardiography video data}

An echocardiography study typically consists of 20--30 ultrasound videos 
   containing multiple views of the heart and vessels with different orientations. 
In our previous study \cite{ulloa2020echo}, we extracted videos with 
   known views.
In the current study, only 28\% of all extracted videos included a description of the source view (e.g. ``AP4'' indicating a four chamber view). 
To overcome this and maximize the number of videos used to train mortality models, 
   we trained view classifier models with the view-labeled data available in the training set. 

From an echocardiography study, we kept only the 
   videos from the parasternal long axis view, the apical two-, four- and five-chamber views, and the basal and mid-ventricular short-axis views.
These views are regarded as the most useful by cardiologists 
    because they capture a large part of the heart's anatomy.
Furthermore, as documented in \cite{ulloa2020echo},
    these views gave the best performance for  
    predicting the risk of one-year mortality, and
    including several millions of videos from additional views
    was deemed infeasible with minimal anticipated benefit.
To  reduce storage requirements,
   the videos were compressed using the default
   Motion-JPEG settings available in OpenCV.
    
\subsection{Electrocardiography trace data}
An ECG test consists of a 10-second recording of 
   electrical potential in the heart for 12 leads at a sampling rate of either 250Hz or 500Hz.
   For this study, we retained the digital voltage data from the 8 non-derived leads (I, II, V1-V6).
For ECG data sampled at 250Hz, we used interpolation to extend
   the dataset to 500Hz.
We extracted the trace data from Geisinger's MUSE (GE Healthcare) 
   and stored it in the lightning memory-mapped database format (\texttt{https://dbdb.io/db/lmdb}). 


\subsection{Training, Validation, and Testing of the Multimodal Model}\label{sec:training}
Starting from all patients with at least an ECG or an echocardiography study (733,245 patients), we randomly assigned 80\% of the patients to the training set, and the rest to the test set.  
All of our results are reported on the test set.

We train the whole multimodal model in three stages.
First, we train six view classifier models (one for each view) in a one-vs-all procedure and collect videos from the desired view to feed the next stage. 
Second, we train the mortality models with convolutional neural network (CNN) architectures (echocardiography videos and ECG traces) independently. The echocardiography video models are trained with 517,348 echocardiography studies found within the training patient list. We train one video model per view with known-view videos and use the model to predict the view of unlabeled-view videos by requiring at least 90\% positive predicted values in the training set. The ECG models are trained with 2,339,622 resting 8-lead ECGs within the training patient list. 
Lastly, we transfer the individually trained CNN models to the proposed multimodal model,  
    where we train again the full model with the 517,348 echocardiography encounters to obtain the feature importance coefficients and the polynomial transformations.
    
For training, we use the Adam optimization method~\cite{kingma2015adam}
    to minimize the binary cross-entropy loss. 
In all training stages, we monitor the training progress with a 
    subset of the training set, i.e the internal validation set. 
The validation set is composed of 
    5\% of the positive cases and an equal number of negative cases, both of which were randomly selected.  
If the internal validation loss does not decrease for more than 10 epochs, 
    we stop the training and recover the model weights at the minimum validation loss~\cite{prechelt1998early}.
    
Here we note that to train the logistic regression and XGB multimodal models, we use the predicted risk scores from each echocardiography video and the ECG traces as input features. However, the proposed model allows us to train using the scalar features along with the echocardiography videos and ECG trace data.

We standardize each scalar factor in the range of $[-1, +1]$.
This is done using:
\begin{equation} \label{eq:minmax}
x_{\text{normalized}}
   = 2\cdot \frac{x - x_{i, min}}{ 
      x_{i, max} - x_{i, min}} - 1,
\end{equation}
where $x_{i, min}$ denotes the minimum value for $x$,
  and $x_{i, max}$ denotes the maximum value for $x$.
We also weigh the error contributions from each class
    based on the number of samples in each class to account
    for sample imbalance.
This is given by     
$$ \textrm{Error weight for i-th class}
    = \frac{N}{NOC*N_i}, $$
where $N$ denotes the total number of samples in all classes,
      $N_i$ denotes the number of samples in class $i$, and
      $NOC=2$ denotes the number of classes.
Here, we use a weight of 4.4 for the positive cases and 0.6 for the negative cases.

To train, we gather all encounters from patients in the training set. 
For testing, we select a random echocardiography encounter 
   per patient to avoid biasing the test results toward patients with more encounters.

We implemented all the experiments with the Tensorflow Docker image \textit{tensorflow/tensorflow:2.8.0-gpu-jupyter} on an NVIDIA DGX-2 platform with 16 V100 32GB GPUs.

\section{Methods}\label{sec:methods}
\begin{figure}[!t]
  \centering
	\includegraphics[width = \columnwidth]{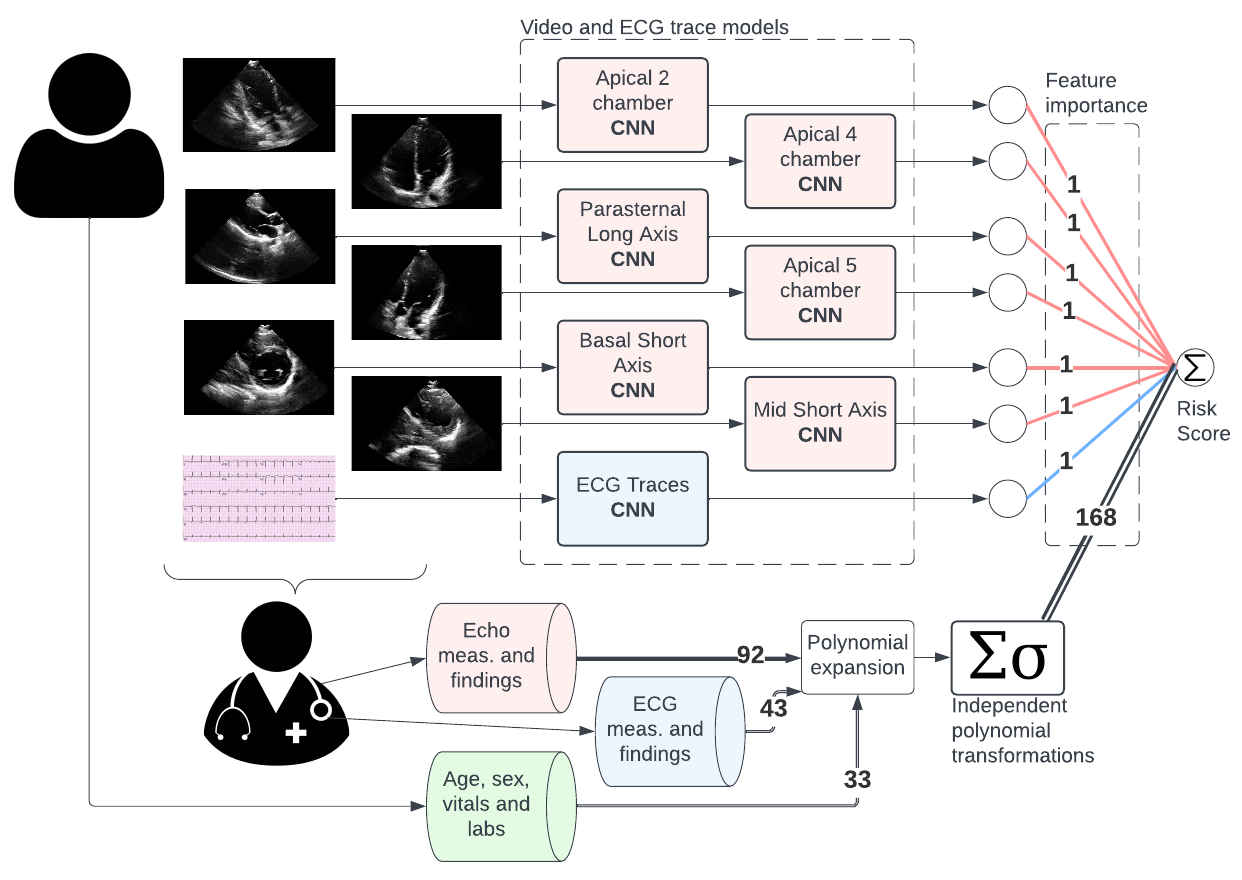}
	\caption{
	      Multimodal system diagram.
	      The system uses echocardiography (`Echo') videos from 6 views (top-left),
	          ECG traces from multiple channels,
	          Echo and ECG measurements and findings,
	          and basic clinical data (age, sex, vitals, and labs).
	      The Echo and ECG measurements and findings are supplied by the physician.
	      3D CNN models are used for processing the videos.
	      A multichannel 1D CNN model is used for training on the ECG traces.
	      The Separable Neural Network system (SNN) is used to process the
	          contributions from the different inputs.
	      The echocardiography videos yield a total of six features (red lines), the ECG traces yield one feature (blue line), and the structured data contribute 168 features (black line). 
	          }
	\label{fig:architecture}
\end{figure}

We present the multimodal processing system in Fig.~\ref{fig:architecture}.
Our system is used to process routine clinical, echocardiography, and ECG data (including multiple videos, traces, measurements, and findings).
    We use CNN models to process the videos and the ECG traces as described below.
We use a separable risk model (SRM) system to combine
    the results from the different modalities to predict one-year mortality.
The SRM produces a risk score that represents the likelihood of mortality within a year.
We note that this retrospective study was approved by the Geisinger Institutional Review Board and performed with a waiver of consent. 
The work falls under the AI in cardiology study (Machine Learning for Data-Driven Precision Medicine) with GIRB\# 2019-0610.

We provide a brief summary of the different modality inputs to the SRM.
We have 33 variables from routine clinical data stored in the EHR.
They include demographics (age, sex, race, smoking status), vitals (body mass index (BMI), blood pressures, etc.), and labs (cholesterol, hemoglobin, etc.).
A list of the 33 variables is given in Table~\ref{tab:varlist}.
We have 92 inputs associated with the measurements and findings from echocardiography
   as given in Table~\ref{tab:varlist}.
The measurements and findings are derived from up to 30 different echocardiography views, but in this study we select six  views
   that include elements of the left ventricle, left atrium, right ventricle, and aortic and mitral valves:
   the parasternal long axis view, the apical two, 
   four and five chamber views, and the basal and mid-ventricular short-axis views.
The videos were aligned to start at the peak of the R-wave of the QRS complex,
    zero-padded or cropped to 2 seconds, sampled at 30 frames per second using
    linear interpolation, and
    downsampled to 109$\times$150 pixels.
In~\cite{ulloa2020echo}, these six echocardiography views were among the best performing for one-year mortality prediction and used a 3D CNN architecture, and we adopt that architecture here. 
The ECG traces were derived from standard clinical 12-lead ECG acquisitions. Specifically, we retained 10-second recordings for each of the 8 non-derived leads,  
    resampled at 500Hz to ensure consistency.
We use a multichannel 1D CNN architecture to process the ECG signal
    as described in~\cite{ulloa2021rechommend}. 
We use 43 ECG-derived measurements and findings, confirmed by
   cardiologists, as described in Table~\ref{tab:varlist}.

The SNN uses a linear combination of the results from
    the different modalities to produce a single score.  
For each one of the continuous variables, we use separable
    polynomial transformations as we will describe below.
For binary variables, the video scores derived from the 3D CNNs,
    and the ECG scores derived from the multichannel 1D CNN,
    we assign a single weight for each one.
Thus, the SNN can be used to directly assess
    the importance of the processed video and ECG data as well as 
    study-derived measurements and findings.

We provide a detailed description of the transformations and the
   multimodal model in section~\ref{sec:multi}.
In section \ref{sec:interp}, 
   we describe how the proposed method supports interpretability
   by ranking features based on their corresponding risks.
Here, we note that the interpretability of the most significant
   features will be further expanded in the results of section 
   \ref{sec:results}.

\subsection{Multimodal Models for Mortality Risk Prediction}\label{sec:multi}
We begin by differentiating categorical factors (e.g., sex), 
      continuous clinical factors (e.g., heart-rate), 
      and risk factors associated with specific modalities (e.g., ECG).
We will construct different models based on the labs, the imaging data, and the ECG.
We will then consider all possible combinations of the different modalities.


We first define all the variables.
We write $X = [x_1, x_2, \dots, x_r]^T$
    for $r$ scalar factors. 
For the 6 views associated with video analysis,
   we have $V = [v_1, v_2, \dots, v_6]^T$.
Then, for the combined outputs for all views, we write
     $M_V = [m_{v1}(v_1), m_{v2}(v_2), \dots, m_{v6}(v_6)]^T$.
We let $m_t$ denote the output associated with the ECG.

We consider a separable model where
    a polynomial transformation is applied to each factor separately.
Thus, the list of all polynomial transformations is given by:
    $$P(X) = [p_1(x_1), p_2(x_2), \dots, p_r(x_r)]^T,$$
    where
    \begin{equation*}
    p(x_i) = \begin{cases} \sigma(a_0 + a_1x_i +a_2x_i^2 + a_3x_i^3 + \dots) &\mbox{if } x_i \in \mathbb{R} \\
x_i & \mbox{if } x_i \in \{0,1\} \end{cases}
    \end{equation*}
    and $\sigma(x) = 1/(1+\exp(-x))$. 
\color{black}     
Here, we are inspired by the use of polynomial regression for linear 
      models to use polynomial transformations to describe
      the non-linear relationship between risk and each input feature.
We note that polynomial transformations require relatively few parameters
      for describing non-linear relationships.
For third-degree polynomials, we need just 4 parameters to
       fit most non-linearities.
As we shall describe in the results, there is some evidence
       that we may not need a higher-order polynomial for modeling risk
       from single features.
In our DISIML package, we allow for a higher degree selection, which can be optimized 
       via hyper-parameter search.
\color{black}

We use a weighted sum of the contributions from each polynomial
    based on:
\begin{equation}
    W_p P(X) = 
          w_1 p_1(x_1)+ w_2 p_2(x_2)+\dots+w_r p_r(x_r),
          \label{eq:poly}
\end{equation}      
where  $W_p = [w_1, w_2, \dots, w_r]$.
Here, we note that the direction of effect in $p_i(x_i)$ could cause the sign of $w$ to flip and provide two solutions, since $\sigma(-x) = 1 - \sigma(x)$.

We thus have that the multimodal risk models are given by:
\begin{align}
    m(X, M_v, m_t) &= \sigma(W\, P_{\text{all}} (X, M_v, m_t) + b) \nonumber \\ 
         &= \sigma(W_p P(X) + W_v M_v + w_t m_t + b)    
         \label{eq:multimodalmodel}
\end{align}
where $ P_{\text{all}} (X, M_v, m_t) = [P(X), M_v, m_t]^T$,
      $W_v = [w_{v1}, w_{v2}, \dots, w_{v6}]$ represents the weights associated with echocardiography views,
      $w_t$ represents the weight associated with the ECG,
      $ W = [W_p, W_v, w_t]$,
      and $b$ represents the bias term. 


\subsection{Separable Models}\label{sec:interp}
Using \eqref{eq:multimodalmodel},
   we compute the final score using:
\begin{align}
  m (X, M_v, m_t) 
    &= \sigma(W\, P_{\text{all}} (X, M_v, m_t) + b) 
       \nonumber \\
    &=  \frac{1}{1 
    + \exp\left( -W\, P_{\text{all}} (X, M_v, m_t)
                 -b\right)}.
  \label{eq:logistic}
\end{align} 
To simplify the notation, 
   let $\pmo = m (X, M_v, m_t)$
   denote the mortality probability.
We note that 
       $\pmo = \sigma(\logit(\pmo))$
       where $\logit(.)=\sigma^{-1} (.)$ represents the inverse of $\sigma(.)$.
We thus have:
\begin{align}
  W\, P_{\text{all}} (X, M_v, m_t)+b & = \logit(\pmo) \nonumber \\
                   & = \log\left(\frac{\pmo}{1-\pmo}\right) 
                     \label{eq:logodds} 
\end{align}    
where $\pmo/(1-\pmo)$ represents the odds ratio. 
Specifically, 
   the product $W\, P_{\text{all}} (X, M_v, m_t) + b$ 
   provides the log-odds of a mortality event \cite{esl}.
Next, to separate out the risk contribution for the $i$-th factor,   
   exponentiate both sides of
eq.~\eqref{eq:logodds} to get:
\begin{equation}
  \text{Odds-ratio}
  = C\cdot \exp(w_i p_i(s)),
    \label{eq:one}
\end{equation} 
where $C$ represents the combined contribution
    from the rest of the features,
   $w_i$ represents the weight from any one of the features, 
   $p_i$ represents any one of the variables, and
   $s$ denotes any one of the contributing variables or modalities.
From eq. \eqref{eq:one},
   it is clear that $w_i$ 
   can be used to understand the contribution
   of the $i$-th feature to the odds ratio
   of the entire model.
In order to quantify the contribution of each feature,
   we rank their corresponding weights as given by:
   $|w|_{(1)} \geq |w|_{(2)} \geq \dots \geq 0$.
To quantify the sensitivity of the risk model for
   a specific feature, we look at 
  the change from $\exp(w_s p_s(s))$  to $\exp(w_s p_s(s+\Delta s))$
  where $\Delta s$ is used to describe a large change in $s$.

\section{Results and Discussion}\label{sec:results}
We begin with a discussion of the most significant features
   in section \ref{sec:feats}.
We then proceed with 
   interpretabilty results for the most significant features 
   in section \ref{sec:riskf}.
We provide two minimal model examples 
    of our proposed separable
   neural network in section \ref{sec:top5}.
We provide comprehensive results for different modalities
   and our multimodality system in section 
   \ref{sec:models}.

\begin{table*}[ht]
  \caption{ \label{tab:coefficients}
  	Top five features with their corresponding coefficients for different modalities. The models were trained based on data aligned with echocardiography encounters.} 
\resizebox{\textwidth}{!}{%
\begin{tabular}{l|lrlrlr|lr}
\toprule
  & \multicolumn{6}{c|}{Single Modality} & \multicolumn{2}{c}{Multimodal} \\ \cmidrule{2-7} \cmidrule{8-9}
  {} & \multicolumn{2}{c}{EHR} & \multicolumn{2}{c}{Echocardiography} & \multicolumn{2}{c|}{ECG} & \multicolumn{2}{c}{EHR+Echocardiography+ECG} \\
  \cmidrule(lr){2-3}\cmidrule(lr){4-5}\cmidrule(lr){6-7}\cmidrule(lr){8-9}
{Rank} & Variable & Coef. & Variable & Coef. & Variable &  Coef. & Variable & Coef. \\
\midrule
1 & Age         &   3.8 & Apical 4 chamber$^*$ &  2.4 & ECG traces  &   4.8 & ECG traces    &  2.3 \\\rowcolor{Gray}
2 & Heart rate  &   2.1 & Apical 2 chamber$^*$ &  1.9 & R axis      &  -0.9 & Age           &  2.1 \\
3 & Hemoglobin  &   1.8 & Severe Tricuspid stenosis &  1.4 & QRS duration &  -0.8 & Apical 4 chamber$^*$ &  1.5 \\\rowcolor{Gray}
4 & eGFR        &   1.5 & Parasternal long axis$^*$ &  1.3 & T Axis &  -0.8 & Apical 2 chamber$^*$ &  1.4 \\
5 & Lymphocytes &  -1.4 & Tricuspid regurgitation max velocity &  1.2 & QTC &  -0.8 & Lymphocytes &  1.4 \\ 
\bottomrule
\multicolumn{9}{l}{$^*$Video data}\\
\end{tabular}
}
\begin{flushleft}
\end{flushleft}
\end{table*}

\subsection{Significant features}\label{sec:feats}
We summarize the results for the most significant features for the
   different models in Table~\ref{tab:coefficients}. 
We begin our discussion with the most significant features
   across modalities.
We then discuss the most significant features
   for each modality in more detail.

As expected for a survival model, 
   age dominates all other features in the basic EHR model. 
For the multimodal model, age remains the 
   second most important feature.
  
The outputs of the CNNs dominate in the echocardiography, ECG, and 
    multimodal model, as opposed to the measurements and findings in both modalities.
For the echocardiography model,
    the outputs of the 3D CNN models associated with the
    two apical views dominate all others.
For the ECG and multimodal models, the output of the 
    multichannel 1D CNN associated with the ECG traces
    is the most significant feature.

Overall, for the echocardiography modality,
      the 3D CNN outputs associated with three views 
     (apical 4 and 2 chambers, and parasternal long-axis) 
     are among the top 5 most important features. 
 The remaining two most significant features include
     the presence of severe tricuspid stenosis (a finding), 
     and tricuspid regurgitation maximum velocity (TRMV, a measurement). 
We recall TRMV measures the maximum velocity of blood flowing backwards from the right ventricle 
   into the right atrium.
The TRMV is an indirect measure of pulmonary artery systolic pressure
    which makes it a marker of pulmonary hypertension. 
The increased mortality risk may be due to pulmonary hypertension
    as discussed in~\cite{samad2018predicting}, 
    which supports clinical interpretability of the model.

For the ECG modality model, the output of the multichannel 1D CNN applied to the ECG
   trace data is the most important feature with 
   a coefficient that is 5 times higher than the second most important.
   The remaining top 5 features comprise standard ECG measurements of features or intervals.
We also note that in~\cite{raghunath2020prediction},
   ECG measurements and findings did not contribute to the model performance as much as the trace data alone. 

For the multimodal model, as mentioned earlier, 
    the output of the ECG trace data was the most significant feature. 
Unfortunately, it is difficult to provide a clinical interpretation of 
    exactly what is being measured by the ECG analysis system.
A previous attempt at interpreting saliency maps in~\cite{raghunath2020prediction} 
    for ECG traces was inconclusive. 
Similarly, for video data, in our previous study~\cite{ulloa2020echo},
    we found that cardiologists could not use the occlusion maps to improve their predictive ability using video data,
    relying on the risk scores instead.
Overall, for mortality risk prediction,
    the dominance of the ECG and video signal processing system in both performance and feature ranks
    suggests that one-year mortality models benefit from incorporating raw data signals. 
Lastly, the polynomial model based on lymphocytes (a lab measurement)
    was found to be as important as 
    the video models associated with the
    apical 4 and 2 chambers views. 
Lymphocytes are part of the immune system and deviations may indicate the 
   presence of a wide range of diseases such as infection, or the presence of pathology within the immune system. 
\begin{figure*}[bht] 
    \centering
    \includegraphics[width=\linewidth]{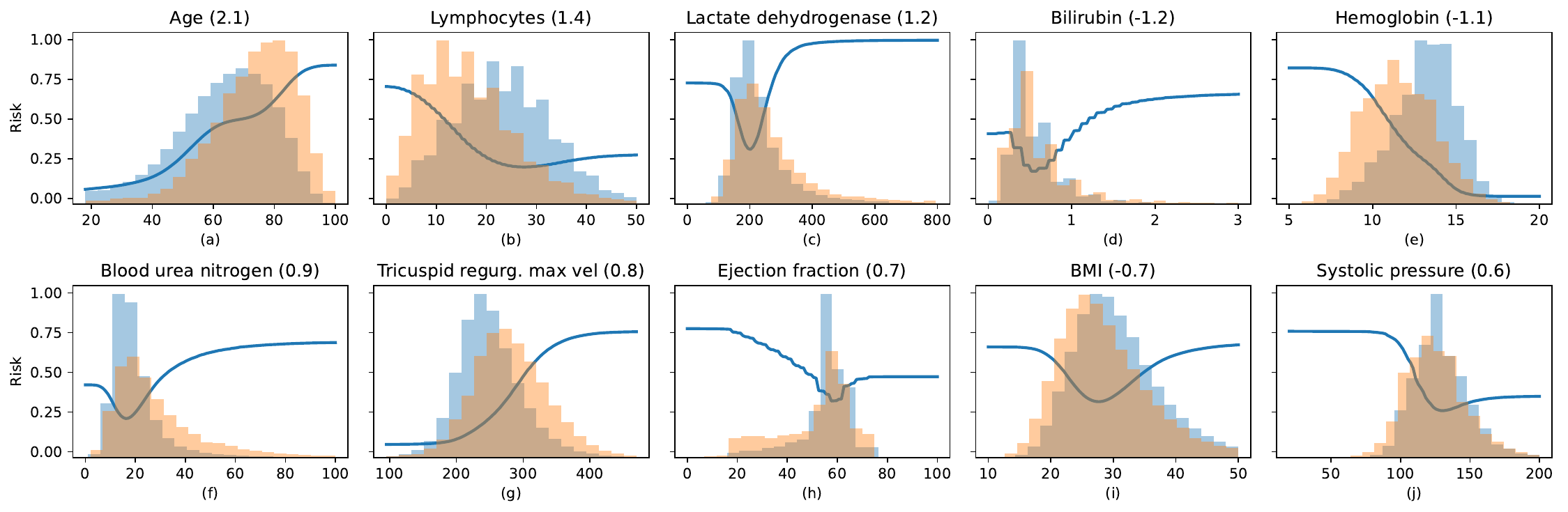}
	\caption{
	      Separable risk functions for the 10 most significant
	           clinical, non-binary features.
	      The risk functions are shown in blue.
	      The normalized histograms of the survivors are shown in light blue.
	      The normalized histograms of the non-survivors appear in orange.
	      When the two histograms overlap, the histograms appear light brown. 
	      Risk functions for: 
	         (a) age in years, 
	         (b) lymphocytes in percent,
	         (c) lactate dehydrogenase in units per liter,
	         (d) bilirubin in milligrams per deciliter, 
	         (e) hemoglobin in grams per deciliter,
	         (f) blood urea nitrogen in milligrams per deciliter,
	         (g) tricuspid regurgitation maximum velocity in centimeters per second,
	         (h) left ventricular ejection fraction in percent,
	         (i) BMI in kilograms per meter squared, and
	         (j) systolic blood pressure in millimeters of mercury.
	     For each risk function, the weight for each feature is given
	         in the title of the graph. If the coefficient is negative, the risk function is corrected to normalize the graph to higher values for higher risk.
	         }
	\label{fig:risks}
\end{figure*}

\subsection{Risk model assessment for individual features}\label{sec:riskf}
In Fig.~\ref{fig:risks}, we show the risk functions associated 
   with the 10 most significant non-binary clinical features
   associated with the multimodal model.
Each risk function represents the
   contribution of $\sigma_s(s)$
   in the multimodal model of eq. 
   \eqref{eq:one}.
Here, we do not show risk functions for the CNNs.
In what follows, we interpret the most significant features.

\color{black}
Based on the risk function diagrams of Fig.~\ref{fig:risks},
    we note that our third-order polynomial model captures 
    some significant non-linearities.
It is clear from the risk trends that a linear model would be inadequate for
    many features.
At the same time, for any given risk level, we never have more than two 
   feature values that map to that risk value.
Here, we note that our cubic model would have allowed up to three feature
    values that map to the same risk value.
Hence, the fact that this never happens provides some evidence that
     we may not need higher order polynomials to capture the trends.            
\color{black}

We begin our analysis with the age factor
   (see Fig.~\ref{fig:risks}(a)). 
We can clearly see mortality increases with age.
Age has a weight coefficient of 2.1 (see Table~\ref{tab:coefficients}).
This implies an 8-fold ($8 \approx e^{2.1}$) increase in the odds ratio between the 
   minimum age of 18 and the maximum age of 98 years old. 

The normal range for lymphocytes is between 20\% and 40\%, which is the range where our analysis shows the risk is the lowest (see Fig. \ref{fig:risks}(b)). Known as lymphocytopenia, decreased lymphocyte levels may indicate a severe infection, such as sepsis, which is the most common cause of death in hospitals in the US~\cite{rudd2020global}.

Following a U-shaped risk trend, lactate dehydrogenase (Fig. \ref{fig:risks}(c)) shows the lowest risk at normal range values, between 105 to 333. Similarly, the normal range for bilirubin (Fig. \ref{fig:risks}(d)) is 0.1 to 1.2.

For hemoglobin, we observed a rapid decline in risk for higher values (Fig. \ref{fig:risks}(e)). This could indicate the higher mortality rate of end-stage anemic patients. We also note that slightly different normal ranges are expected for males and females. For future iterations of this model we could add a hemoglobin and sex interaction term to assess whether the interaction coefficient is of significant importance. 

The risk trend for blood urea nitrogen (Fig. \ref{fig:risks}(f)) was also U-shaped with the lowest risk at normal range values, 6 to 24.

We have a strong, positive trend for increases in the TRMV (see Fig. \ref{fig:risks}(g)). 
Based on our prior discussion on the TRMV, 
  this trend is to be expected and compatible with pulmonary
  hypertension being strongly linked to mortality.

The model yielded a U-shaped trend for the left ventricular EF (see Fig.~\ref{fig:risks}(h)). 
This trend agrees with the findings described in~\cite{wehner2019routinely}.
It is important to recognize that the majority of the patients fall
   within the non-linear trend region and thus it would be a false assumption
   to suggest a linear trend for the entire EF region. Moreover, the performance drop from 0.76 (using the proposed model) to 0.73 (using logistic regression), shown in the first row of Table \ref{tab:aucs_tab}, also suggests that non-linearity improves performance.
We recall that EF denotes the percentage of blood that leaves the left ventricle chamber
   during contraction.
Based on the current American Heart Association guidelines (reviewed as of May 31, 2017),
    normal EF values are between 50\% and 70\%.
We can clearly see that our model correctly modeled this range as low-risk 
    (see Fig.~\ref{fig:risks}(h)).
On the other hand, high EF may be a marker of a hyperdynamic heart failure
   with preserved ejection fraction~\cite{wehner2019routinely}.
Yet, for heart failure, low EF is well recognized as a poor prognosticator.

The model also yielded a U-shaped trend for BMI with an optimal value between 25 and 30 (see Fig.~\ref{fig:risks}(i)). While current guidelines from the World Health Organization, define a normal BMI range as 18.5 to 24.9, a recent study~\cite{winter2014bmi} suggests that for adults older than 65 BMIs lower than 23 or higher than 33 were the most harmful. Geisinger's clinical population has a median age of 66, which aligns better with~\cite{winter2014bmi}.

We note that there is a notable increase in risk for
   low values of the systolic pressure (see Fig.~\ref{fig:risks}(j)). 
There are two possible interpretations for this trend.
First, we note that high blood pressure
    may not lead to 1-year mortality.
Instead, high blood pressure results in long-term 
   cumulative effects such as renal 
   and heart failure which results in an increased risk for longer-term mortality.
Second, we note that low blood pressure may be a marker of cardiac decompensation.
Here, we note that cardiac decompensation is more acutely associated with hospitalization and death. 
Clearly, in order to fully understand the trend, further study will be needed,
    taking into account how patient medications affect blood pressure.

\begin{table}[!b]
  \caption{
  Summary of minimum and maximum values
     used for  normalizing features for
     the minimal models (see eq. \eqref{eq:minmax}).}
  \label{tab:norm}
  \centering
  \begin{tabular}{llrr}
\toprule
{} &        Units &Min &        Max \\
    \midrule
Age                         & years         & 18    &  98 \\
Heart Rate (HR)             & beats per minute           & 40    & 150 \\
BMI                         & kg/m$^2$      & 14.4  & 76.3 \\
Diastolic Pressure (DP)     & mmHg          & 32    & 120 \\
Systolic Pressure (SP)      & mmHg          & 72    & 213 \\
Lymphocytes (Lym)           & percent     & 1     & 91 \\
Lactate Dehydrogenase (LDH) & units/L     & 74    & 2513\\
Bilirubin (Bil)             & mg/dL    & 0.1   & 10.4 \\
Hemoglobin (Hgb)            & g/dL          & 6.3   & 19\\
\bottomrule
\end{tabular}
\end{table}

\subsection{Minimal Models}
\label{sec:top5}
We also consider two minimal models that avoid any 
    features that are based on ECG or echocardiography exams.
For the first minimal model,
    we consider common variables that can be 
    monitored at home with a weight scale and a blood
    pressure monitor: 
    age, heart rate (HR), BMI, systolic (SP) 
    and diastolic (DP) blood pressures.
For the second minimal model, 
    we selected the top 5 scalar features from the multimodal model 
    that were not ECG or echocardiography based features: 
    age, lactate dehydrogenease (LDH), bilirubin (Bil), 
    lymphocytes (Lym), and hemoglobin (Hgb).
As we did for all other models, we standardize each factor
    in the range of -1 to +1.
For completeness, we provide the min and max values in Table
    \ref{tab:norm}.

The mortality risk based on the first minimal model gave:
\begin{align}
  \text{Risk}&= \sigma\, \bigl( \nonumber\\
  & 3.62  \, \sigma( 1.82 \cdot \text{Age}^{3} - 0.53 \cdot \text{Age}^{2} + 0.66 \cdot \text{Age} -0.09) \nonumber\\
  &+ 2.81 \, \sigma( 1.53 \cdot \text{HR}^{3}  + 0.17 \cdot \text{HR}^{2}  + 0.52 \cdot \text{HR}  -1.35) \nonumber\\
  &+1.60 \, \sigma(  -0.42 \cdot \text{BMI}^{3} + 1.96 \cdot \text{BMI}^{2} - 0.30 \cdot \text{BMI} -1.48) \nonumber\\
  &+1.37 \, \sigma(  -2.98 \cdot \text{DP}^{3}  - 1.26 \cdot \text{DP}^{2}  - 1.42 \cdot \text{DP}  -1.68) \nonumber\\
  &+2.75 \, \sigma(  -0.34 \cdot \text{SP}^{3}  + 0.86 \cdot \text{SP}^{2}  - 0.09 \cdot \text{SP}  +0.21) \nonumber\\
  &-5.09\bigr). \tag{7} \label{eq:minimal1}
\end{align}
In eq. \eqref{eq:minimal1}, we have normalized the expression
   to give positive weights for the final layer using 
   $\sigma(-x) = 1 - \sigma(x)$.
Based on the last layer coefficients,
  we have the features ranked in the order of:
  age, heart rate, systolic pressure, BMI, and diastolic pressure.
However, the most exciting result is that
   this simple model gave an AUC of 0.78, which
   approximates the performance of far more
   sophisticated models discussed in section
   \ref{sec:models}.

The mortality risk for the second model gave:
\begin{align}
    \text{Risk} &= \sigma\, \bigl( \nonumber\\
    &3.76 \, \sigma( 1.95  \cdot \text{Age}^{3} - 0.80 \cdot \text{Age}^{2} + 0.59 \cdot \text{Age} - 0.71) \nonumber\\
    &+2.40\, \sigma( -1.38 \cdot \text{Lym}^{3} - 0.75 \cdot \text{Lym}^{2} - 2.03 \cdot \text{Lym} - 1.58) \nonumber\\
    &+1.33\, \sigma( -2.33 \cdot \text{LDH}^{3} - 6.18 \cdot \text{LDH}^{2} - 2.18 \cdot \text{LDH} + 0.93) \nonumber\\
    &+2.38\, \sigma( 1.47 \cdot  \text{Bil}^{3} + 1.22 \cdot \text{Bil}^{2} - 0.52 \cdot \text{Bil} - 1.96) \nonumber\\
    &+2.27\, \sigma( 3.13 \cdot  \text{Hgb}^{3} + 1.56 \cdot \text{Hgb}^{2} + 1.84 \cdot \text{Hgb} + 1.33) \nonumber\\
    &-3.40\bigr). \tag{8} \label{eq:minimal2}
\end{align}
In eq. \eqref{eq:minimal2},
  we have the features ranked in the order of:
  age, lymphocytes (Lym), bilirubin (Bil), hemoglobin (Hb), and lactate dehydrogenease (LDH).
As expected, this simplified model performed slightly better
   than the first minimal model by giving an AUC of 0.8.

From both equations \eqref{eq:minimal1} and \eqref{eq:minimal2}, 
    we can see that the non-linear effects are very significant. 
The non-linear coefficients for the second and third-order terms are in some cases  
    higher than the linear terms and cannot be captured by a linear regression model.
In many ways, our approach can be considered as a generalization of linear logistic regression.
It is impressive that our minimal models of  \eqref{eq:minimal1} and \eqref{eq:minimal2}
   are based on just 5 contributing factors.

\begin{table}
  \centering
  \caption{
  Summary statistics on all echocardiography studies, where the patient survived or died within a year of the study. We show the study and patient counts; features averages, standard deviations (in parenthesis) and interquartile ranges (in brackets).}
  \label{tab:demo}
  \resizebox{\columnwidth}{!}{%

  
  \begin{tabular}{l|rr|rr}
  \toprule
   & \multicolumn{2}{c|}{\textbf{Alive}} & \multicolumn{2}{c}{\textbf{Deceased}} \\ \midrule
\# Echo studies & \multicolumn{2}{c|}{539,689} & \multicolumn{2}{c}{68,049}\\ \rowcolor{Gray}
 \# Patients & \multicolumn{2}{c|}{248,250} &  \multicolumn{2}{c}{46,031} \\ \midrule
   & \textbf{Mean (SD)} & \textbf{IQR} & \textbf{Mean (SD)} & \textbf{IQR}  \\ \midrule
Age                 & 64 (16) & [54, 75] & 73 (14) & [65, 84] \\ \rowcolor{Gray}
Heart Rate          & 75 (15) & [64, 83] & 80 (17) & [68, 89] \\
BMI                 & 31 (8)  & [26, 35] & 29  (9) & [24, 33] \\ \rowcolor{Gray}
Diastolic Press.    & 73 (12) & [64, 80] & 68 (13) & [60, 76] \\
Systolic Press.     & 129 (19) & [117, 140] & 125 (22) & [110, 139] \\ \rowcolor{Gray}
LDH                 & 251 (266) & [173, 263] & 320 (387) & [189, 325]\\
Hemoglobin          & 15 (45) & [12, 14] & 12 (33) & [10, 13] \\ \rowcolor{Gray}
Bilirubin           & 0.56 (0.58) & [0.30, 0.70] & 0.81 (1.64) & [0.30, 0.80] \\
Lymphocytes         & 23 (10) & [16, 30] & 17 (12) & [9, 22] \\
\bottomrule
  \end{tabular}

  }
\end{table}

\subsection{Multimodality Model results}\label{sec:models}
We provide comprehensive results from a large 
   number of models
   in Table \ref{tab:aucs_tab}.
Here, we note that the models demanded
   significant computational resources.
Even with 16 V100 32GB GPUs,
   it took us 1.5-2 months to train
   the different models on 
   the large datasets 
   described in section \ref{sec:dataset}.
In Table \ref{tab:aucs_tab}, we include results from
   logistic regression, the separable neural network (SNN),
   and XGB. We set the XGB model with a learning rate of 0.05, with 300 estimators, max depth of 4, and a subsample of 80\%.
We note that logistic regression
   is a special case of the SNN, where we use
   linear instead of cubic polynomials to describe
   the trends.

We show our use of low-parameter
    models in Table \ref{tab:aucs_tab}.
For example, the 3D CNN model for video analysis
     only required 14,309 parameters.
Similarly, the 1D CNN model for processing all the
     ECG traces only required 18,561 parameters.
The combined model that uses all of EHR, 
     Echo, and ECG only required 33,398 parameters.
Our minimal models require just 25 or 20 parameters.
Throughout our table, we can see
      a clear trade-off between
       lower accuracy achieved by minimal models (AUC: 0.76 and 0.78)
       and
       higher accuracy achieved by low-parameter unbiased
       multimodal models (max AUC: 0.88).

Unfortunately, we note that it is not possible
    to compare our low-parameter models to
    high-parameter models due to the fact
    that high-parameter models would 
    require prohibitively larger training
    times.
   

We begin with the analysis of the results from
   individual modalities.
We will then consider combinations
   of two or more modalities.

In terms of single modality models, 
     echocardiography gave the best AUC at 0.86,
     closely followed by EHR at 0.85,
     and electrocardiography at 0.81.
For echocardiography,
    video analysis significantly outperformed
    predictions that were solely based
    on measurements and findings.
For EHR, 
    the inclusion of all features
    significantly outperformed the use of age
    and sex alone.
For ECG,    
    the multichannel CNN model gave
    the best results that did not
    improve with the inclusion of 
    ECG measurements and findings. Similar to echocardiography, the ECG CNN outperformed the ECG measurements and findings model, and the combination did not improve compared to the CNN. This suggests that the CNNs are finding important features in both echocardiography and ECG that we are not measured clinically.

The minimal models of equations
    \eqref{eq:minimal1} and \eqref{eq:minimal2}
    perform below the single modality
    models.
Nevertheless, with AUCs of 0.78 and 0.80,
    they rival the performance of 
    the electrocardiography model at 0.81.
However, as mentioned earlier,
    the minimal models are far easier to understand
    and they can be implemented with simple
    at-home measurements or using 4 common
    blood panel measurements.

The multimodality models outperformed
    the single modality models.
In the full model, 
    the combination of all modalities
    gave the best performance with an AUC of 0.90.
The full model was only slightly better
    than EHR+echocardiography (0.89)
    or EHR+electrocardiography (0.88).
In terms of performance, 
    XGB performed slightly better
    than SNN and logistic regression.
However, in most cases, SNN performed almost at
    the same level as XGB.
For the full model, SNN
    gave an AUC of 0.89 (compared to 0.90 for XGB).
    
The results from the single and multimodal models also
    outperformed our prior results on smaller datasets.
In \cite{samad2018predicting}, 
    the combination of echocardiography measurements and findings, and EHR  studies
    using a random forest and 331,317 echocardiography gave an AUC of 0.85.
In the current paper, the corresponding
   model would be EHR all + Echo m. that gave 
   better AUC values at 0.86 with SNN, and 0.87 with XGB.

The multimodal model significantly outperformed a much smaller study described
    in ~\cite{ulloa2020echo}.
In~\cite{ulloa2020echo}, the combination of echocardiography measurements and findings, 
    video analysis from 24 views, and EHR data with 42,095 studies gave an AUC of 0.84. 
In the current paper, the corresponding model would be 
  EHR all + Echo all that gave 0.88 with SNN and 0.89 with XGB.
Here we note that we include six of the 24 views because of significant demands
  on compute time constraints. 
A model that includes all 24 views has the potential to further improve the overall performance.
\begin{table}
  \centering
  \caption{ Model performances in percent AUC measured at a randomly selected Echocardiography study per patient. LR stands for Logistic regression; SNN for 
  Separable Neural Network, the proposed method; and XGB for Extreme Gradient Boosting.}
  \label{tab:aucs_tab}

\begin{tabular}{p{4cm}llll}
\toprule
 & SNN & LR  &  SNN & XGB \\
 & Params & AUC & AUC  & AUC \\
\midrule
\multicolumn{4}{l}{ \textbf{Echocardiography}}\\ 
Echo measurements and findings & 296   &  0.73 &  0.76 &  0.78 \\ 
Videos using 3D CNNs           & 85854 &  0.84 &  0.84 &  0.86 \\
Echo all                       & 86150 &  0.85 &  0.85 &  0.86 \\[0.05in]
\multicolumn{4}{l}{ \textbf{Electrocardiography}}\\ 
ECG measurements and findings  & 79    &  0.73 &  0.75 &  0.76 \\
ECG traces using 1D CNN        & 18561 &  0.81 &  0.81 &  0.81 \\
ECG all      & 18640 & 0.81 &  0.81 &  0.81 \\[0.05in]
\multicolumn{4}{l}{\textbf{EHR}}\\ 
Age + Sex                 & 10  &  0.72 &  0.72 &  0.72 \\
Age + Sex + EHR (EHR all) & 153 &  0.82 &  0.84 &  0.85 \\[0.05in]
\multicolumn{4}{l}{\textbf{EHR + Echocardiography}}\\ 
Echo m. + Age + Sex  & 306   &  0.78 &  0.80 &  0.81 \\
Videos + Age + Sex   & 85864 &  0.85 &  0.85 &  0.86 \\
Echo all + Age + Sex & 86160 &  0.85 &  0.86 &  0.87 \\
EHR all + Echo m.    & 449 &  0.84 &  0.86 &  0.87 \\
EHR all + Videos     & 86007   &  0.87 &  0.88 &  0.89 \\
EHR all + Echo all   & 86303  &  0.87 &  0.88 &  0.89 \\[0.05in]
\multicolumn{4}{l}{\textbf{EHR + Electrocardiography}}\\ 
Age + Sex + ECG m.   & 89    &  0.79 &  0.81 &  0.81 \\
Age + Sex + ECG tr.  & 18571 &  0.83 &  0.83 &  0.83 \\
Age + Sex + ECG all. & 18650 &  0.83 &  0.84 &  0.84 \\
EHR all + ECG m.     & 232   &  0.84 &  0.86 &  0.87 \\
EHR all + ECG tr.    & 18714 &  0.86 &  0.87 &  0.88 \\
EHR all + ECG all    & 18793 &  0.86 &  0.87 &  0.88 \\[0.05in]
\multicolumn{4}{l}{\textbf{EHR + Echocardiography + Electrocardiography}}\\ 
Age + Sex + videos + ECG tr.  & 104425 &  0.87 &  0.87 &  0.88 \\
EHR all + videos + ECG tr.    & 104568 &  0.88 &  0.89 &  0.90 \\
EHR all + Echo m. + ECG m.    & 528   &  0.85 &  0.87 &  0.87 \\
EHR all + Echo all + ECG m.   & 86382 &  0.88 &  0.89 &  0.89 \\
EHR all + Echo all + ECG tr.  & 104864 &  0.88 &  0.89 &  0.90 \\
EHR all + Echo all + ECG all* & 104943 &  0.88 &  0.89 &  0.90 \\[0.05in]
\multicolumn{4}{l}{\textbf{Minimal Models}}\\
Age, Heart Rate, BMI, and Blood Pr        & 25 &  0.76 &  0.78 &  0.78 \\ 
Age,  LDH, Bilirubin, Lymphocytes and Hb  & 20 &  0.78 &  0.80 &  0.80 \\
\bottomrule
\multicolumn{4}{l}{* Full model}
\end{tabular}

\end{table}

In summary, the current paper introduces
   the use of separable neural
   networks to process a significantly
   larger dataset than was previously
   considered.
The results for both single-modality and 
   multimodality models have been significantly improved.    
In addition, the proposed separable neural network
   can be used to provide meaningful interpretations
   for different factors and across modalities.
We note that the relationship between 
   the dominant clinical factors and the overall risk is
   highly non-linear (see Fig. \ref{fig:risks}).
In addition, we note that these non-linear relationships
   cannot be effectively captured by linear regression models.

\section{Conclusion}\label{sec:concl}
The paper describes the development of 
    separable models for risk assessment based on
    a massive multimodal dataset.
The paper also develops the methodology
     that allows us to quantify the contribution
     of each clinical factor to the overall risk.
For predicting all-cause 1-year mortality,
    the developed methods 
    produced significantly better performance
    over prior models that used ECG, echocardiography, and EHR.
The implementation code
    is provided in
    the DISIML package~\cite{ulloa2019disiml}.

Our models have been validated 
    on the largest dataset of its kind.
For development and validation,
    the paper derives models based on 
    25,137,015 videos 
        associated with 699,822 echocardiography studies from 316,125 patients, 
        and 2,922,990 8-lead ECG traces from 631,353 patients.
\color{black}        
We also study the effectiveness
       of 84 low-parameter models.
For all models, we support separable risk assessment
       that allows us to associate risk based on individual features.
Furthermore, our separable models enable us to perform
       optimal feature selection based on the weights
       associated with each feature in the global model.       
The paper introduces a minimal model 
       based on Age, Heart Rate, BMI, and Blood Pressure
       that can be monitored at home.
At just 25 parameters, this minimal model achieves
       an AUC of 0.76.        
Based on optimal feature selection,
       we introduce an optimized minimal model based on
       Age,  LDH, Bilirubin, Lymphocytes, and Hb.
At just 20 parameters, our optimized minimal model
       achieves an AUC of 0.78.       
At an AUC of 0.78, the optimized minimal model
         performs better than the  
         use of echo measurements and findings
         (AUC=0.76 with 296 parameters) and
         ECG measurements and findings
         (AUC=0.75 with 79 parameters).
Nevertheless, our low-parameter 3D CNN 
         performed significantly better
         (AUC=0.84 with ~86k parameters).
Overall, we have found
        that using additional information
        will improve performance.
For the full multimodal model,
        with nearly 105k parameters,
        we have achieved an AUC of 0.89,
        a substantial improvement over our optimized 
        minimal model.
\color{black}
\balance
\bibliographystyle{IEEEtran}  
\bibliography{main}

\end{document}